\pdfoutput=1
\documentclass[11pt,dvipsnames]{article}

\usepackage{acl}

\usepackage{dsfont}
\usepackage{times}
\usepackage{latexsym}
\usepackage{color}
\usepackage{multirow}
\usepackage{forest}
\usepackage{tikz}
\usepackage{algorithm}
\usepackage{algpseudocode}
\usepackage{amsmath}
\usepackage{graphics}
\usepackage{graphicx}
\usepackage{subfig}
\usepackage{mwe}
\usepackage{amssymb}
\usepackage{bm}
\usepackage{amsmath}
\usepackage{makecell}
\usepackage{tabularx}
\usepackage{arydshln}
\usepackage{url}
\usepackage{CJKutf8}
\usepackage{wrapfig}
\usepackage{enumitem}
\usepackage{calc}

\newlength\myheight
\newlength\mydepth
\settototalheight\myheight{Xygp}
\usepackage[T1]{fontenc}
\usepackage[utf8]{inputenc}

\usepackage{CJKutf8}
\usepackage{booktabs}
\usepackage{multirow}
\usepackage{graphics}

\usepackage{tabularx}
\usepackage{diagbox}
\usepackage{xcolor}
\usepackage{array}

\usepackage{microtype}

\usepackage{inconsolata}

\title{OMGEval\includegraphics[width=1.1em]{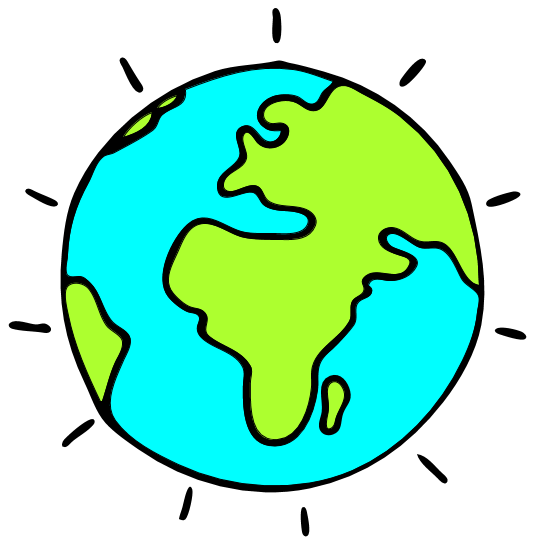}: An Open Multilingual Generative Evaluation Benchmark\\for Large Language Models}

\author{Yang Liu$^{1}$ \
  Meng Xu$^{1}$ \
  Shuo Wang\thanks{Corresponding authors.}$^{2}$ \
  Liner Yang$^{*1}$ \\
  \textbf{Haoyu Wang}$^{3}$ \
  \textbf{Zhenghao Liu}$^{4}$ \
  \textbf{Cunliang Kong}$^{2}$ \
  \textbf{Yun Chen}$^{5}$ \\
  \textbf{Yang Liu}$^{2}$ \
  \textbf{Maosong Sun}$^{2}$ \
  \textbf{Erhong Yang}$^{1}$ \\
  $^1$Beijing Language and Culture University \quad $^2$Tsinghua University \\
  $^3$Beijing University of Posts and Telecommunications \quad $^4$Northeastern University \\
  $^5$Shanghai University of Finance and Economics\\
  }

\begin{document}
\begin{CJK*}{UTF8}{gbsn}

\maketitle
\begin{abstract}

Modern large language models (LLMs) should generally benefit individuals from various cultural backgrounds around the world. However, most recent advanced generative evaluation benchmarks tailed for LLMs mainly focus on English. To this end, we introduce OMGEval, the first \textbf{O}pen-source \textbf{M}ultilingual \textbf{G}enerative test set that can assess the capability of LLMs in different languages. For each language, OMGEval provides 804 open-ended questions, covering a wide range of important capabilities of LLMs, such as general knowledge, logical reasoning, and so on. Each question is rigorously verified by human annotators. Notably, to sufficiently reflect the compatibility of LLMs in different cultural backgrounds, we perform localization for each non-English language. Specifically, the current version of OMGEval includes 5 languages (i.e., Zh, Ru, Fr, Es, Ar). Following AlpacaEval, we employ GPT-4 as the adjudicator to automatically score different model outputs, which is shown closely related to human evaluation. We evaluate several representative multilingual LLMs on the proposed OMGEval, which we believe will provide a valuable reference for the community to further understand and improve the multilingual capability of LLMs. 
OMGEval is available at \url{https://github.com/blcuicall/OMGEval}.
\end{abstract}

\section{Introduction}
\begin{figure}[ht]
    \centering
    \includegraphics[width=1\linewidth]{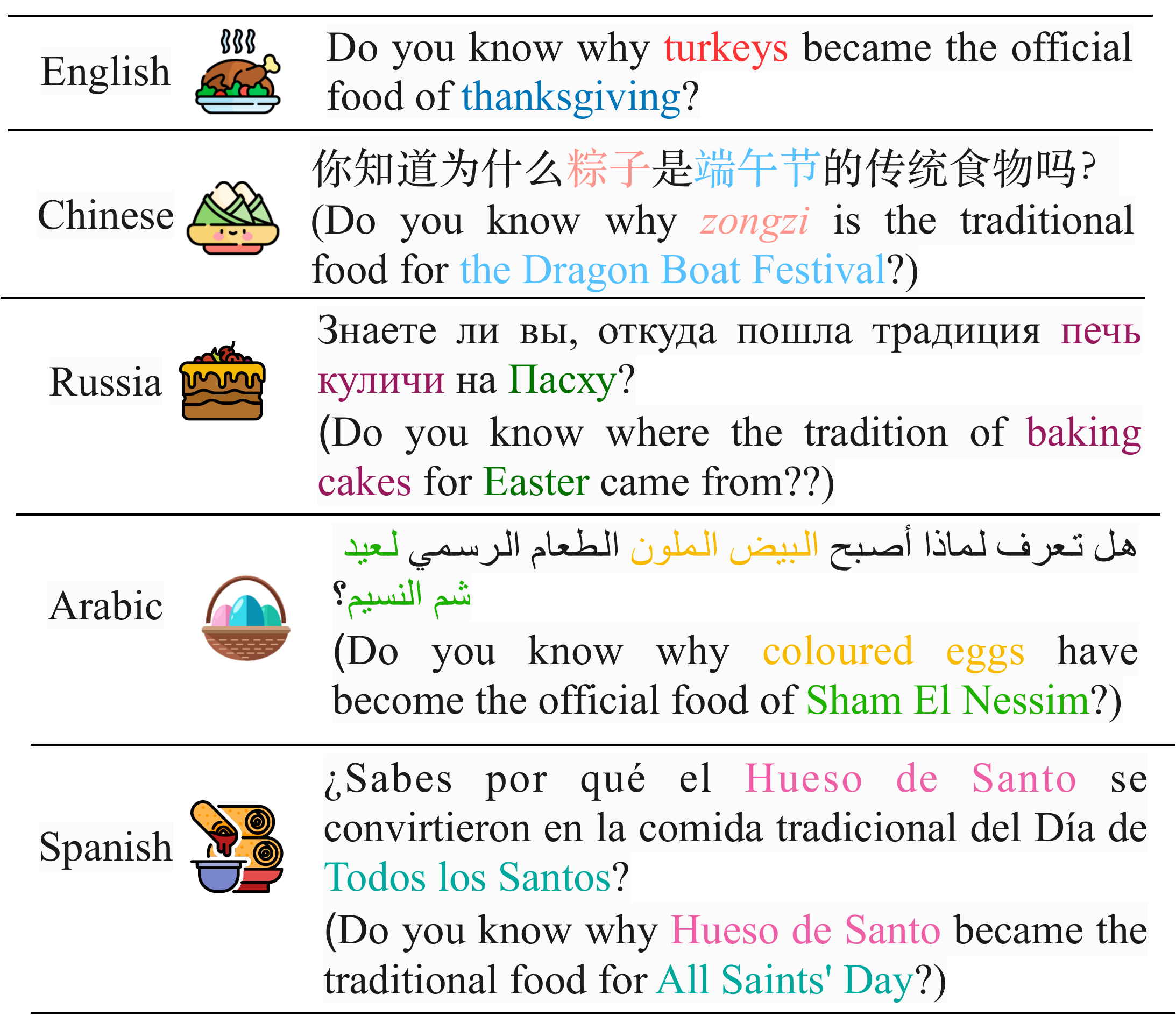}
    \caption{Example for question localization, the language-specific items are highlighted in different colors. In different cultural contexts, discussions about the same topic can vary significantly. For instance, when talking about festivals and food, Americans might focus on Thanksgiving and turkey, while Chinese people may discuss the Dragon Boat Festival and Zongzi.}
    \label{fig:example-food}
\end{figure}

Large Language Models (LLMs) have recently demonstrated
remarkable capabilities in natural language processing tasks and beyond~\cite{Naveed2023ACO,openai2023gpt4}. More than 7,000 languages\footnote{\url{https://www.ethnologue.com}} are spoken around the world nowadays, with a considerable number facing the challenges of being low-resourced, under-represented, or disappearing~\cite{singh2024aya,gao2023measures}. In contrast, the most widely used datasets and breakthroughs in LLMs have coalesced in English. Enhancing the multilingual capabilities of LLMs is crucial for benefiting a wider audience, allowing more people to enjoy the advantages of large model applications~\cite{workshop2022bloom,wei2023polylm,lai2023okapi}.

\begin{figure*}[ht]
    \centering
    \includegraphics[width=0.99\linewidth]{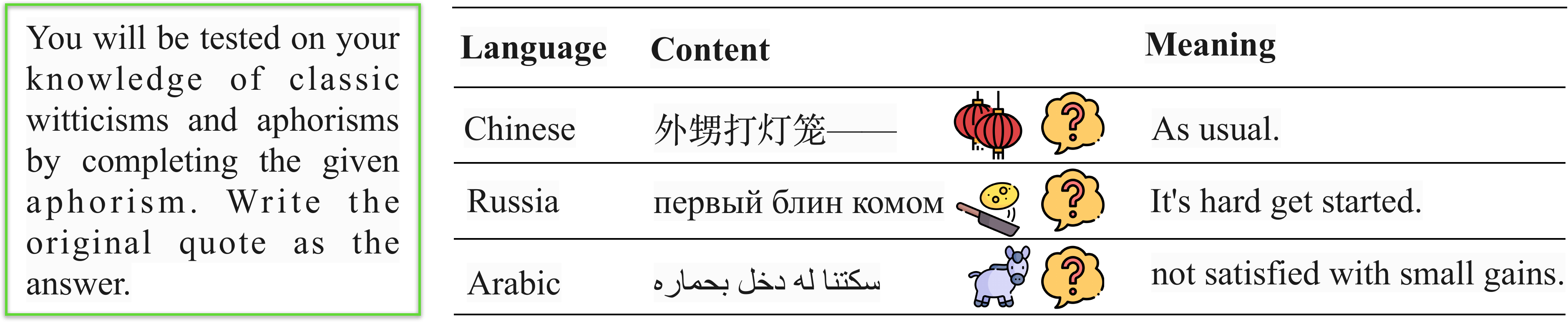}
    \caption{An question that requires LLMs to complete the proverb given the prefix. Proverbs in different languages are diverse and may be difficult to understand without the knowledge of the corresponding language.}
    \label{fig:example3}
\end{figure*}

However, evaluations of the multilingual capabilities of LLMs are currently inadequate~\cite{Yuan2023HowMI}. Most previous multilingual benchmarks mainly focus on discriminative tasks, which are not specifically tailored for LLMs~\cite{conneau2019unsupervised,yang2019paws,ponti2020xcopa,bandarkar2023belebele}. Recently, some English generative benchmarks~\cite{alpaca_eval,zheng2023judging} are proposed, which are shown more effective in reflecting the capability of LLMs.
% benchmarks like AlpacaEval~\cite{alpaca_eval} and MT-Bench~\cite{zheng2023judging} have adopted open-ended question answering for evaluation.
However, these modern benchmarks primarily lie in English, leaving the generative evaluation of multilingual LLMs rather unexplored.
% resulting in a limited understanding of LLMs’ capabilities in other languages\cite{huang2023ceval}. 

In this work, we proposed \textbf{OMGEval}, an \textbf{O}pen \textbf{M}ultilingual \textbf{G}enerative \textbf{Eval}uation benchmark for LLMs, which recognizes and considers the richness of diverse cultural backgrounds around the world. The advantages of OMGEval include:

\begin{itemize}
    \item {\em Localized Questions}: as the necessary communication medium among people, language can reflect cultural diversity across various countries and regions, containing both general and culture-specific information. In different cultural contexts, discussions about the same topic can vary significantly. As illustrated in Table \ref{fig:example-food}, people speaking different languages may focus on distinct items even under the same topic.
    % when talking about festivals and food, Americans might mention Thanksgiving and turkey, while Chinese individuals are more likely to refer to the Dragon Boat Festival and {\em zongzi}.
    To this end, we construct sufficient localized questions that can better reflect the cultural backgrounds of different languages.
    During the localization process, phrases in questions that involve cultural references, such as names of people, places, festivals, and foods, are adapted to fit the cultural context of the target language.
    \item  {\em Rigorous human verification}: to ensure the reliability of OMGEval, each question is polished and verified by human experts in different languages.
\end{itemize}

We conduct experiments to evaluate several representative multilingual LLMs on OMGEval. The results show that GPT-4 is the only model that surpasses the 50 average win rate. However, its 55.52 win rate indicates that OMGEval is a challenging benchmark for current LLMs. In contrast, the performance of other open-source multilingual models reveals a significant gap in the ability to process and understand cultural nuances, indicating a broader issue within the field. This disparity underscores the critical need for a concerted effort in the community to address cultural biases and enhance the global applicability of LLMs. 
% Future research should focus on refining these models for greater cultural sensitivity and developing methodologies for more equitable data representation and model training processes.

\section{Backgroud}
\subsection{Importance of Multilingo Evaluation}
\paragraph{Applicability and Generalizability}
Language usage varies significantly across different social and cultural groups ~\cite{trudgill2000sociolinguistics}, highlighting the need for datasets that capture various linguistic expressions and cultural contexts. This confirms that we evaluate language models in a manner that mirrors their real-world applicability across diverse user groups.Figure~\ref{fig:example3} presents idioms from Chinese, Russian, and Arabic cultures. Individuals unfamiliar with a particular culture often find it challenging to grasp the intended meaning. Incorporating folk sayings into language models serves as an effective measure for evaluating their ability to navigate subtle cross-cultural differences.

\begin{figure*}[ht]
    \centering
    \includegraphics[width=0.99\linewidth]{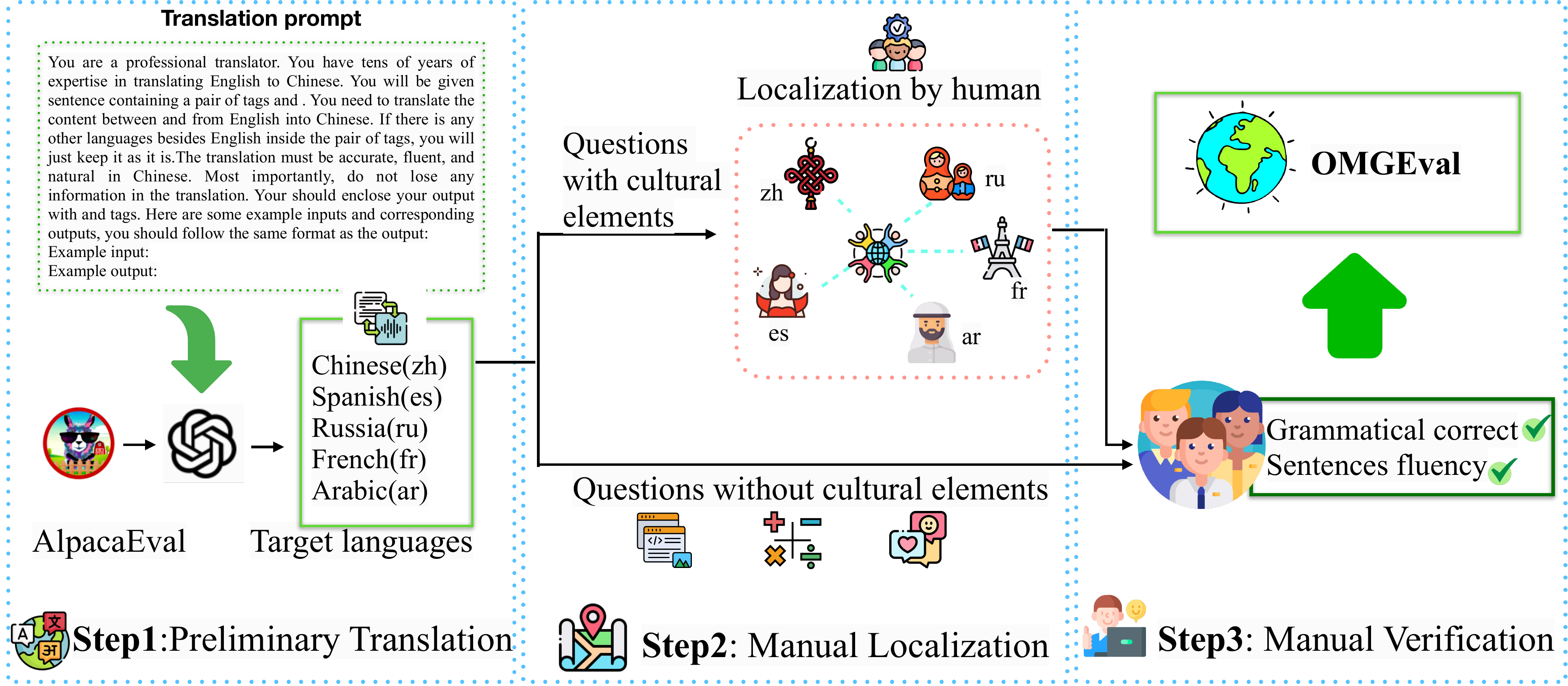}
    \caption{Construction process of OMGEval.}
    \label{fig:overview}
\end{figure*}

\paragraph{Combating Cultural Hegemony} English benchmarks tend to exhibit geographical biases towards the domestic knowledge of the regions that produce them ~\cite{huang2023ceval}. \citet{gramsci2020selections}points to the suppression of minority or marginalized cultures by dominant cultural norms and values. Language models limited to monolingual or monocultural datasets have been identified within the Ethical AI domain as prone to developing biases~\cite{Talboy2023ChallengingTA,Thakur2023UnveilingGB,Tao2023AuditingAM}. If a dataset only includes holidays mainly celebrated in the United States, such as Thanksgiving, it may unintentionally prioritize and spread specific cultural norms and values.

\paragraph{Real-World Scenario Simulation}

The meaning of language is derived from its use in real-life situations~\cite{rabiah2018language}. Thus, when evaluating LLMs, questions that would naturally occur in everyday situationsit is essential. For example,Thanksgiving is almost never mentioned in the context of Chinese people. Traditional multilingual benchmarks for NLP tasks, often derived by translating English datasets into other languages, have been criticized for introducing an English-centric bias ~\cite{liu2021visually}. This occurs because the translation process may need to encapsulate culturally specific concepts or nuances inherent to the target languages. The predominance of English in existing datasets overlooks the gnored the real-life situations of people speaking other languages, particularly those considered low-resource~\cite{singh2024aya}.  
\begin{table*}[ht]
    \centering
    \small
    \begin{tabular}{cl}
      \toprule
      \multirow{1}{*}{\bf Categories} & {\bf Examples in OMGEval} \\
       \midrule
       \multirow{3}{*} {\shortstack {{\includegraphics[height=1.9\myheight]{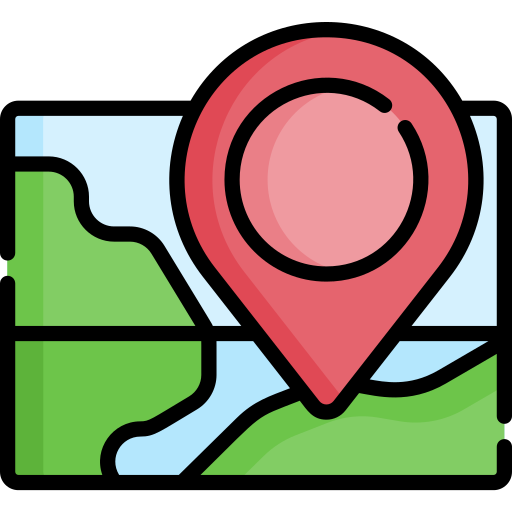}} \\
      \bf Name of Places}} & \textbf{before:} How did \textcolor{red}{US states} get their names? \\
      & \textbf{after:} \textcolor{red}{中国各个省份}的名字是怎么来的？ \\
      & \hspace{0.8cm} {\tiny ({\em How did \textcolor{red}{the provinces of China} get their names?})} \\
      \midrule
      \multirow{3}{*}{\shortstack {{\includegraphics[height=1.9\myheight]{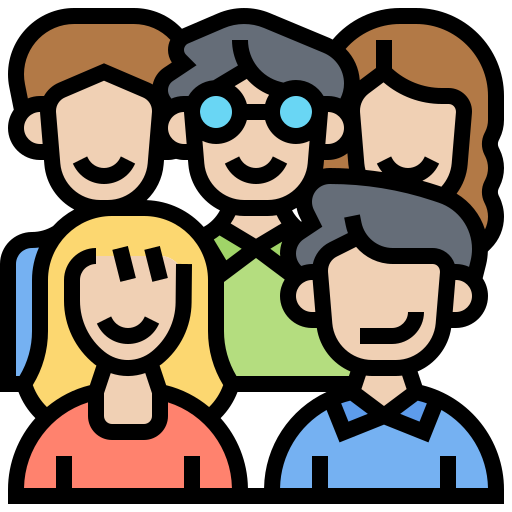}} \\
      \bf Name of the person}} & \textbf{before:} Who is \textcolor{red}{Larry Page}?\\
      & \textbf{after:} \textcolor{red}{马化腾}是谁？\\
      & \hspace{0.8cm} {\tiny ({\em Who is \textcolor{red}{Ma Huateng}?})} \\
      \midrule
      \multirow{3}{*}{\shortstack {{\includegraphics[height=1.9\myheight]{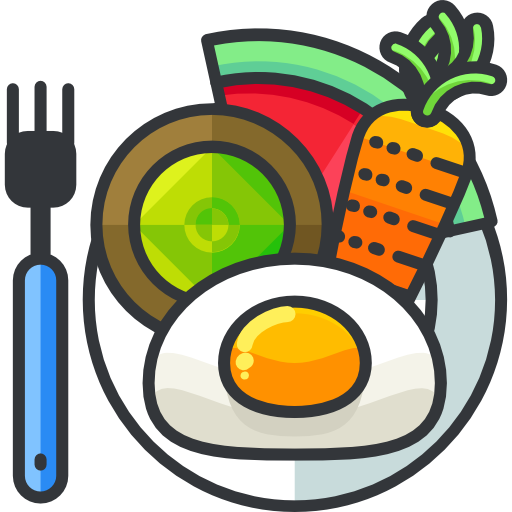}} \\
      \bf Food}} & \textbf{before:} Do you know why \textcolor{red}{turkeys} became the official food of thanksgiving? \\
      & \textbf{after:} 你知道为什么\textcolor{red}{粽子}是端午节的传统食物吗？\\
      & \hspace{0.8cm} {\tiny ({\em Do you know why \textcolor{red}{zongzi} is the traditional food for the Dragon Boat Festival})} \\
      \midrule
      \multirow{3}{*}{\shortstack {{{\includegraphics[height=1.9\myheight]{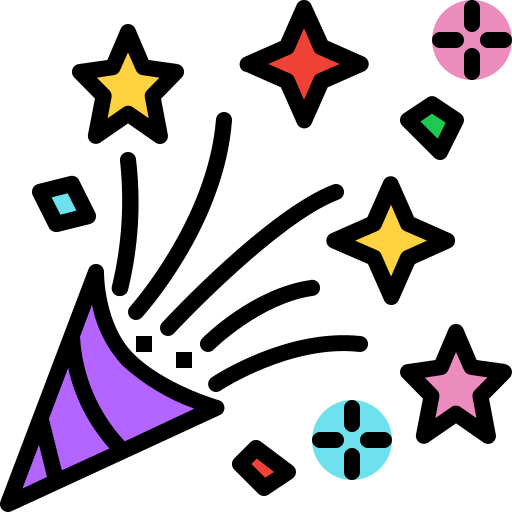}}} \\
      \bf Festival}} & \textbf{before:} Are there any weird \textcolor{red}{Christmas} traditions? \\
      & \textbf{after:} \textcolor{red}{春节}有什么特别的习俗吗？\\
      & \hspace{0.8cm} {\tiny ({\em Are there any special customs for \textcolor{red}{Chinese New Year}?})} \\
      \midrule
      \multirow{3}{*}{\shortstack {{{\includegraphics[height=1.9\myheight]{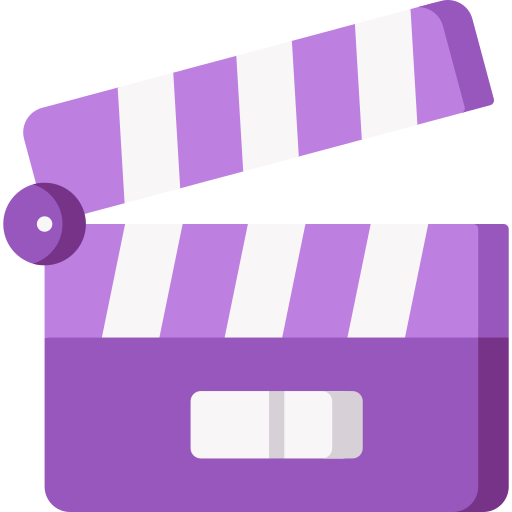}}}} 
      \bf \shortstack{TV show、movies\\books、games}} & \textbf{before:} who does lady gaga play in \textcolor{red}{american horror story}? \\
      & \textbf{after:} \textcolor{red}{《霸王别姬》}中张国荣扮演谁？\\
      & \hspace{0.8cm} {\tiny ({\em Who does Leslie Cheung play in \textcolor{red}{Farewell My Concubine}?})} \\
      \midrule
      \multirow{3}{*}{\shortstack {{{\includegraphics[height=1.9\myheight]{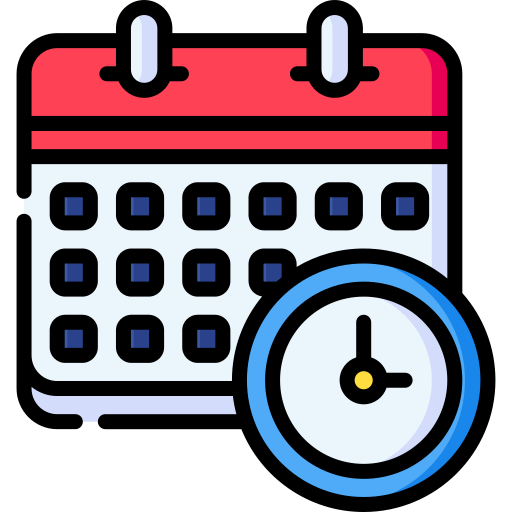}}} \\
      \bf Period}}  & \textbf{before:} What if the Internet had been invented during \textcolor{red}{the Renaissance period}? \\
      & \textbf{after:} 如果\textcolor{red}{明朝}就有了互联网会怎样？\\
      & \hspace{0.8cm} {\tiny ({\em What if the internet had existed in \textcolor{red}{the Ming Dynasty}?})} \\
      \midrule
      \multirow{3}{*}{\shortstack {{{\includegraphics[height=1.9\myheight]{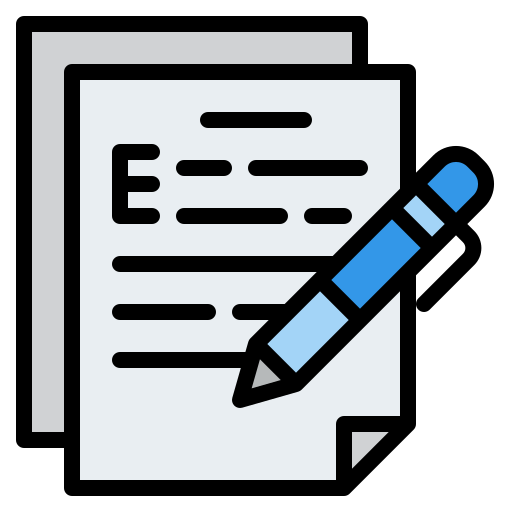}}}}
      \bf \shortstack{Language and \\writing system}}& 
      \textbf{before:} Identify all words that match the pattern given \textcolor{red}{H\_AR\_}.\\
      & \textbf{after:} 找出尽可能多的符合给定格式的词语：\textcolor{red}{AABB式}。\\
      & \hspace{0.8cm} {\tiny ({\em Find as many idioms as possible that fit the given format: \textcolor{red}{AABB style}.})} \\
      \bottomrule
    \end{tabular}
    \caption{Some representative topics that often contain cultural elements. 
    % The localization process takes into consideration, but is not limited to, the above seven points.
    }
    \label{tab:localization}
\end{table*}
\subsection{Necessity of Generative Evaluation}

\paragraph{Complexity and Range of Outputs}
LLMs have exhibited remarkable capabilities to understand, reason, and generate texts across a variety of open-ended tasks~\cite{chung2022scaling,openai2023gpt4,openai_chatgpt}. However, current LLMs mainly evaluate their performance previous benchmarks that are not tailored for evaluating LLMs with open-form output~\cite{liang2023holistic,huang2023ceval,arora2023llms,clark2018think}. For example, in MMLU~\cite{hendrycks2021measuring}, an answer is considered correct only if the model’s output exactly matches the ground truth answer. These benchmarks can only measure LLMs’ core capability on a confined set of tasks (e.g. multi-choice knowledge or retrieval questions), which fails to assess their alignment with human preference in open-ended tasks adequately~\cite{vicuna2023,li2023prd,nakano2022webgpt,ning2024peerreviewinllms}.
Generative evaluation allows models to demonstrate their ability to produce coherent, contextually appropriate, and diverse outputs, which can comprehensively evaluate the capabilities exhibited by the model.

\section{Data Collection}

We aim to construct a multilingual, open-ended Q\&A dataset, which can comprehensively evaluate LLMs' capabilities. Leveraging the widely acknowledged AlpacaEval ~\cite{dubois2023alpacafarm}, we have curated 805 entries as our foundational data. In the subsequent sections, we will explain the procedures for preliminary multilingual translation, manual localization, and the rigorous manual verification of this raw data to ensure its efficacy and global relevance, as shown in Figure~\ref{fig:overview}.

\subsection{Preliminary Translation}

Our dataset is aligned across all languages, ensuring equal examples are available in every language. To translate raw English data into other languages, we employ GPT-4 ~\cite{achiam2023gpt}, utilizing specific prompts tailored for each language to guarantee accuracy and contextual relevance in the translations. The primary purpose of translation is to provide a reference for localization and manual calibration so that those involved in this work can understand or rewrite the sentences efficiently.

\subsection{Manual Localization}

Not all data requires localization; we focus on adapting culturally specific elements, such as names of people, places, or traditions, as shown in Table ~\ref{tab:localization}. Precise translations are deemed adequate for sentences without cultural elements like code writing and mathematical problem-solving. To accomplish localization effectively, annotators must deeply understand English culture and the target language's culture. During the localization, it is crucial to maintain consistency across multiple facets, such as people, events, and timelines. For instance, \textit{What if the Black Death had not occurred in the 14th century?} Not only do we need to map \textit{the Black Death} to \textit{the Great Plague} (大瘟疫), but also to contextualize \textit{the 14th century} with the specific historical time of \textit{the Ming Dynasty} (明朝). Furthermore, efforts should be made to preserve as much feature similarity as possible when localizing cultural content. Take \textit{Hawaii} for example, we should change it to \textit{SanYa} (三亚) rather than \textit{Mount Tai} (泰山), both \textit{Hawaii} and \textit{SanYa} are renowned holiday destinations with closer similarities in their appeal as tourist spots. This approach stresses the importance of retaining cultural content's similarities and guarantees that the localized data suits the target audience well, enhancing the relevance and relatability of the information.

\subsection{Manual Verification}
The annotation team assigned to each language comprises two annotators and one reviewer, each possessing a master's degree in the linguistics of the respective language. The criteria for the manual calibration include the construction of grammatically correct and fluent sentences alongside the demonstration of logical coherence.

\subsection{Evaluation}
\begin{figure}[ht]
    \centering
    \includegraphics[width=0.99\linewidth]{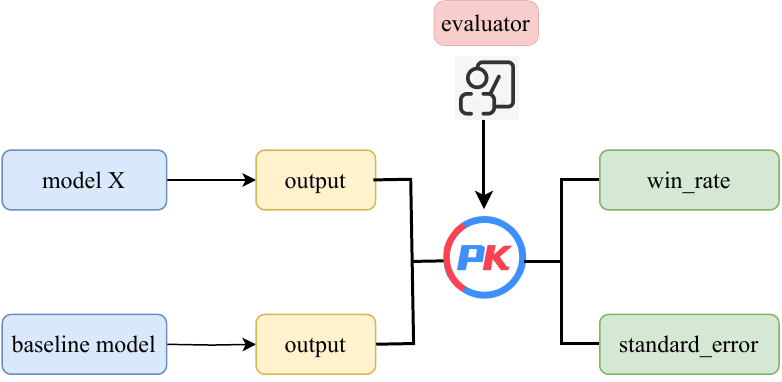}
    \caption{Illustration of the evaluation process.}
    \label{fig:evaluation}
\end{figure}
AlpacaEval ~\cite{alpaca_eval} stands as a comprehensive benchmark for the automated evaluation of LLMs. It utilizes a dataset comprising 805 instructions and amalgamating evaluation data from various projects, including Self-instruct ~\cite{wang2022self}, Open Assistant, and Vicuna ~\cite{chiang2023vicuna}. The evaluation process, as shown in Figure~\ref{fig:evaluation}, employs a high-capacity model, typically GPT-4, to serve as the adjudicator, which automatically gauges the responses of the baseline model and the model to-be evaluated, by comparing win rates. The experimental results from AlpacaEval revealed a Pearson correlation coefficient ~\cite{cohen2009pearson} of 0.93\% between GPT-4 and human-judged results, underscoring the evaluation's high reliability. 

Following AlpacaEval, we adopt a similar evaluation strategy, selecting a model's output as a baseline and employing GPT-4 as the adjudicator. This process involves comparing the evaluated model's output with baseline and calculating the win rate and standard error. To guarantee the model's outputs align with the intended language we expect, we specifically tailor the prompts to include a question in the target language, such as utilizing a Chinese question prompt to evaluate Chinese subset.

\begin{table*}[ht]
    \centering
    \small
    \begin{tabular}{l m{0.65\linewidth}}
    \toprule
    \bf Type & \bf Content \\
    \midrule
    {\includegraphics[height=1.2\myheight]{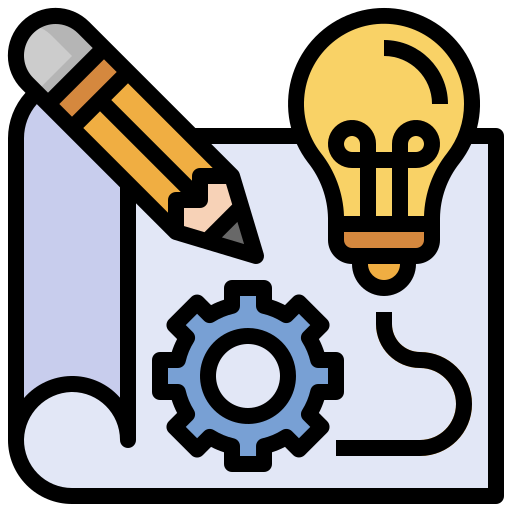}} Generation and Creation & Marketing proposals, creating advertisements, style writing, etc. \\
    \cmidrule(lr){1-1} \cmidrule(lr){2-2}
    \includegraphics[height=1.2\myheight]{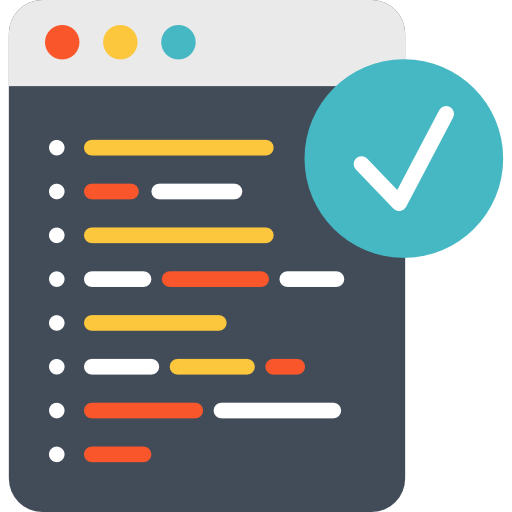}
    Language Comprehension & Grammar checking, reading comprehension, information extraction, contextual dialogue, etc. \\
    \cmidrule(lr){1-1} \cmidrule(lr){2-2}
    \includegraphics[height=1.2\myheight]{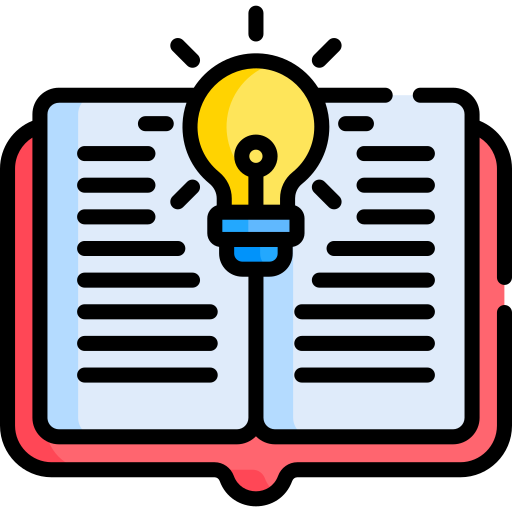}
    General Knowledge & Less specialized, life-like knowledge quiz\\
    \cmidrule(lr){1-1} \cmidrule(lr){2-2}
    \includegraphics[height=1.2\myheight]{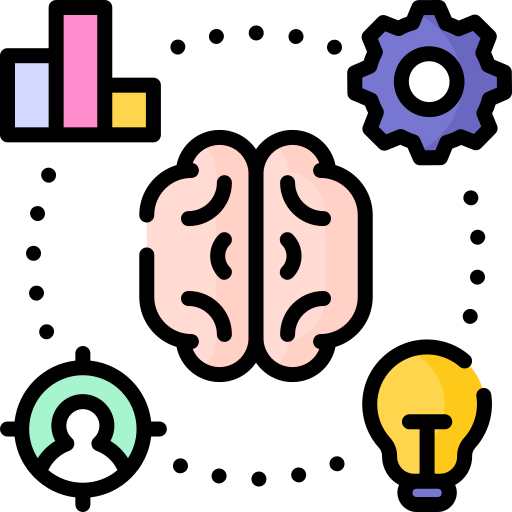}
    Professional Knowledge & More specialised knowledge quiz\\
    \cmidrule(lr){1-1} \cmidrule(lr){2-2}
    \includegraphics[height=1.2\myheight]{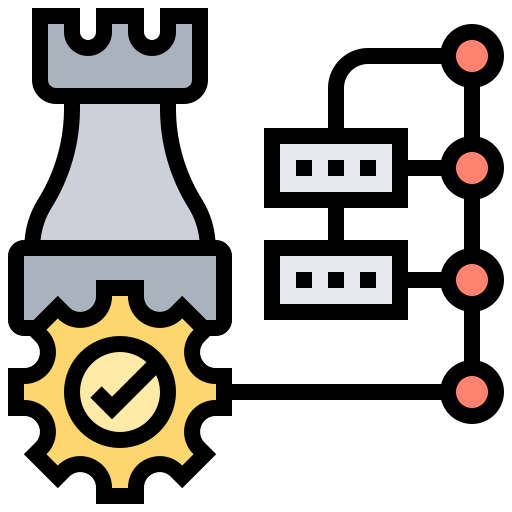}
    Logical Reasoning & Common sense reasoning, scientific reasoning, humanistic reasoning, etc. \\
    \cmidrule(lr){1-1} \cmidrule(lr){2-2}
     \includegraphics[height=1.2\myheight]{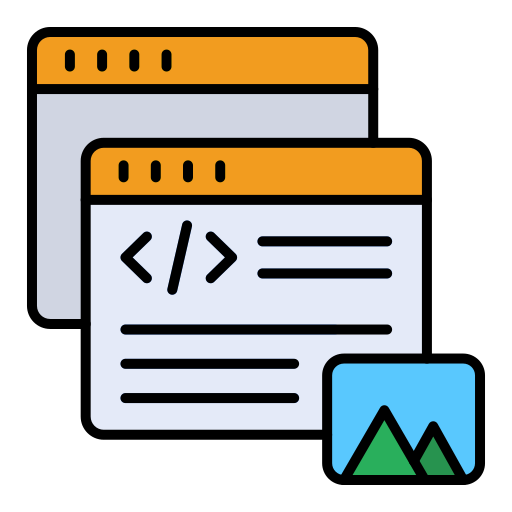}
    Code Skills & Code comprehension, code generation, code modification, etc. \\
    \cmidrule(lr){1-1} \cmidrule(lr){2-2}
     \includegraphics[height=1.2\myheight]{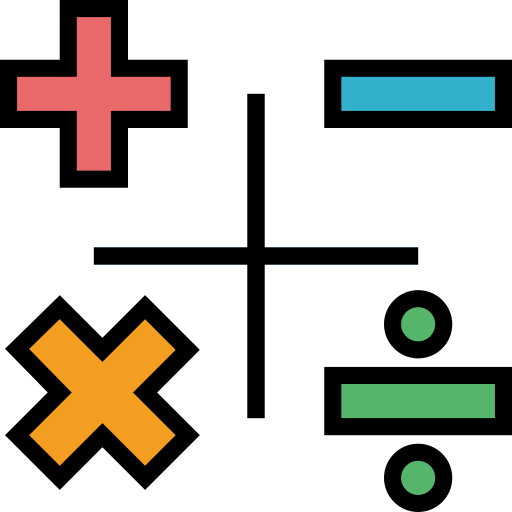}
    Maths Competence & Calculation, algebra, geometry, solving equations, etc. \\
    \cmidrule(lr){1-1} \cmidrule(lr){2-2}
    \includegraphics[height=1.2\myheight]{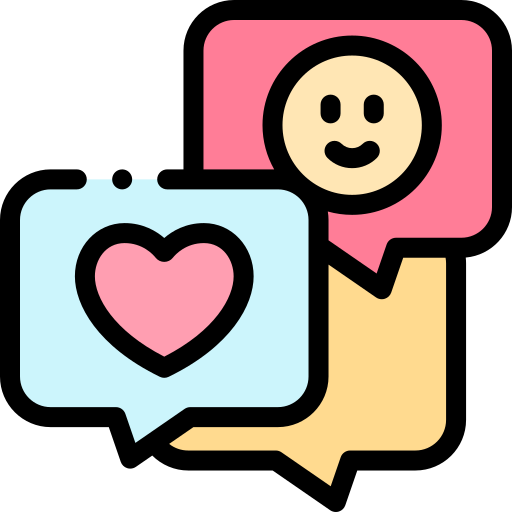}
    Chit chat & Greeting, aimless conversation \\
    \cmidrule(lr){1-1} \cmidrule(lr){2-2}
    \includegraphics[height=1.2\myheight]{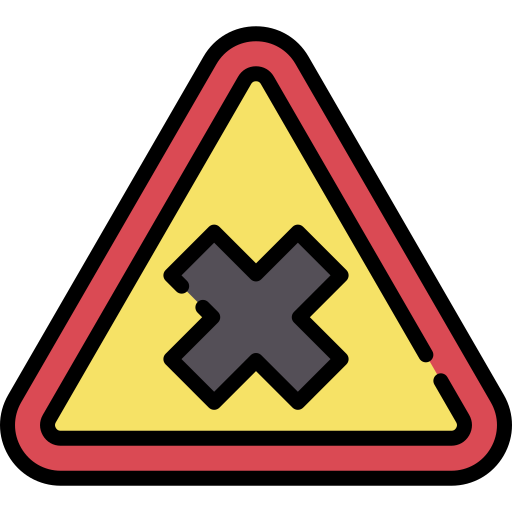}
    Harmlessness & Actions involving religion, discrimination, breaking the law, etc. \\
    \bottomrule
    \end{tabular}
    \caption{9 capability types in OMGEval and their specific contents}
    \label{tab:capability}
\end{table*}

\subsection{Data Analysis}
\paragraph{Capability Type}
\begin{figure}[ht]
    \centering
    \includegraphics[width=0.99\linewidth]{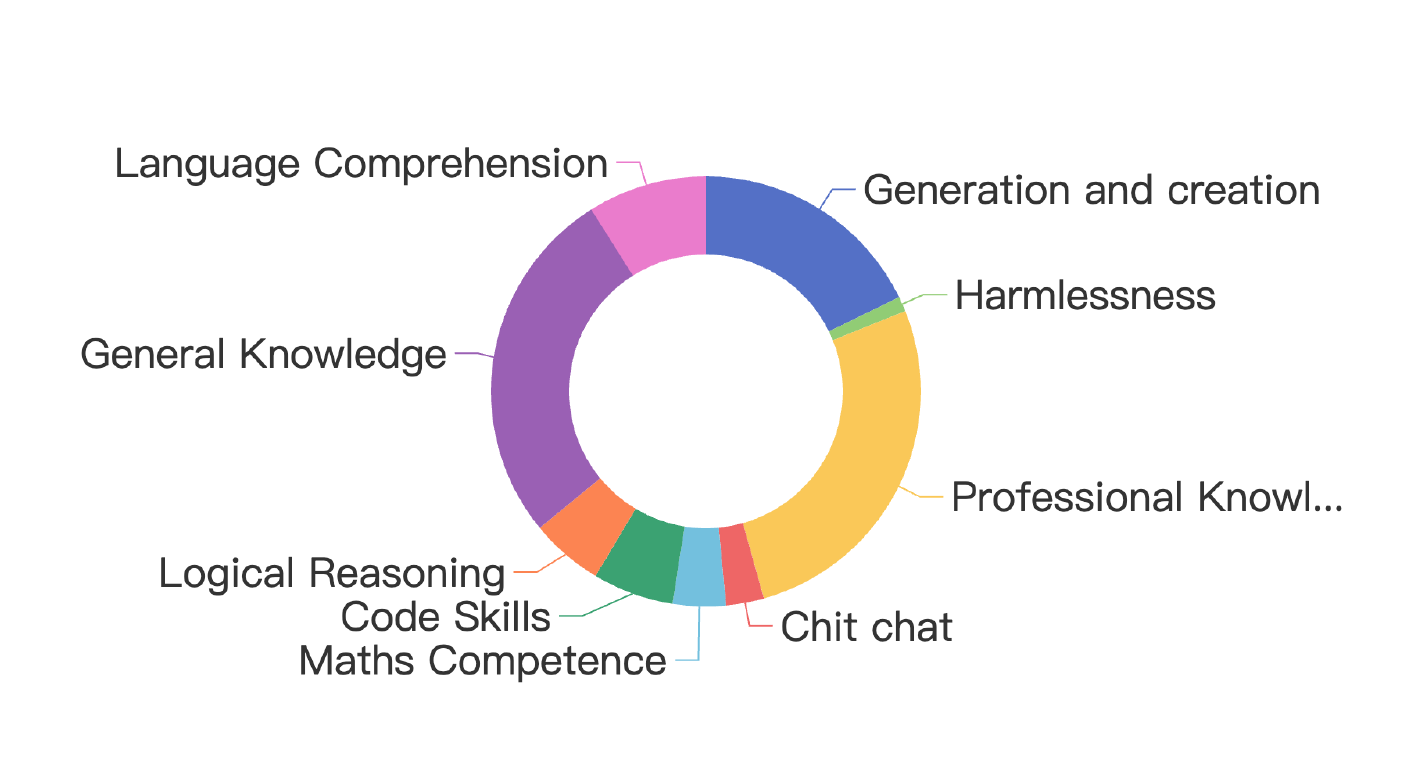}
    \caption{Distribution of questions in OMGEval: General Knowledge (27.0\%), Professional Knowledge (26.7\%), Generation and creation (17.8\%), Language Comprehension (9.0\%), Code Skills (6.1\%), Logical Reasoning (5.5\%), Maths Competence (4.0\%), Chit chat (2.9\%), Harmlessness (1.1\%).}
    \label{fig:catagory}
\end{figure}
In our study, we have categorized the evaluation of abilities into nine distinct types. Table~\ref{tab:capability} shows the detailed tasks corresponding to each type. Each question within our dataset has been meticulously annotated by humans, culminating in the ability assessment distribution depicted in Figure~\ref{fig:catagory}, which shows that the current dataset's question distribution across the various abilities is uneven. The areas of \em{General Knowledge} and \em{Professional Knowledge} predominate, while \em{Harmlessness} and \em{Chit Chat} are less represented. Acknowledging this discrepancy, we will introduce additional open-ended questions to the dataset in the future, enhancing the robustness and diversity of our ability evaluation.

\paragraph{Localization Subset}
\begin{table}[ht]
    \centering
    \small
    \begin{tabular}{l cc}
      \toprule
      \bf Language & \bf Full & \bf Local. Subset \\
      \midrule
      Chinese (Zh) & 804 & 231 \\ 
      Russia (Ru) & 804 & 181 \\
      French (Fr) & 804 & 197 \\
      Spanish (Es) & 804 & 197 \\
      Arabic (Ar) & 804 & 212 \\
      \bottomrule
    \end{tabular}
    \caption{Number of questions for each language. We report the numbers of both the full test set and the localization subset.}
    \label{tab:subset_analyse}
\end{table}

Each language features a distinct subset of localized data, with varying quantities reflecting the cultural differences from English. The extent of localization required increases with the cultural divergence from English. The more a culture differs from English, the more localization is needed. As illustrated in Table~\ref{tab:subset_analyse}, Chinese and Arabic have undergone the most extensive localization, highlighting the significant cultural differences between English-speaking regions and others in the real world.

\begin{figure}[ht]
    \centering
    \includegraphics[width=0.99\linewidth]{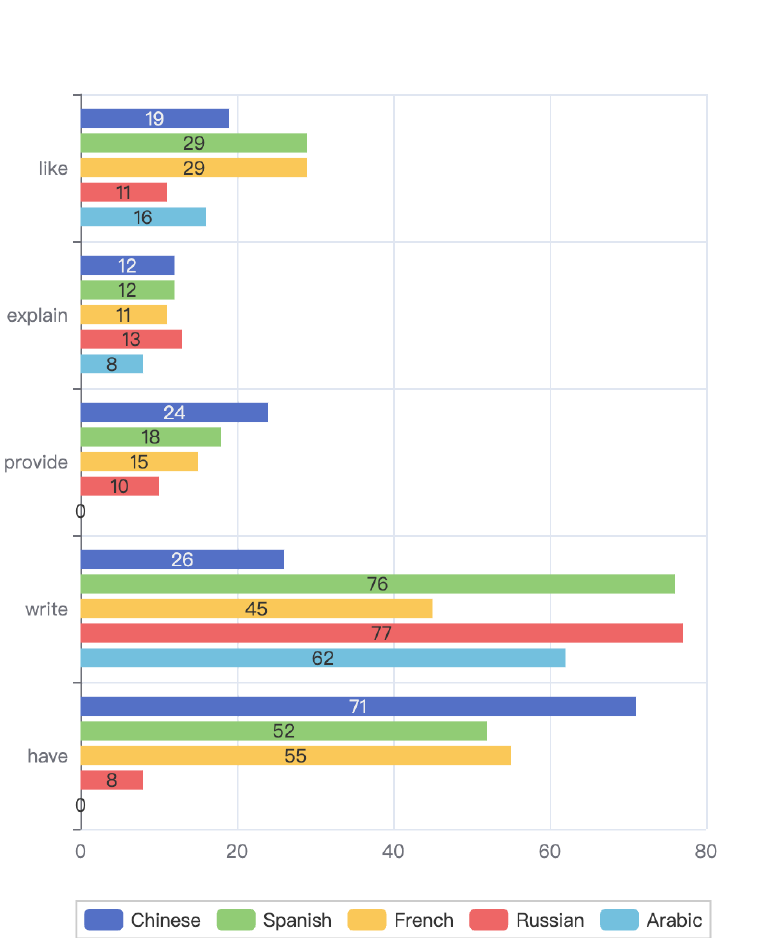}
    \caption{5 frequent root verbs of OMGEval dataset.}
    \label{fig:root_verb}
\end{figure}
\paragraph{Root Verb Distribution}
Root verbs can manifest different intentions and patterns within a dataset directive. We conducted root verb analysis on sentences in OMGEval. Figure ~\ref{fig:root_verb} illustrates five verbs that frequently appear across five languages. Variations in expression habits across languages lead to differences in the distribution of root verbs. For instance, in the Chinese dataset, the term "write" can be expressed as "编写"(write), "写"(write), "撰写"(write), "起草"(write), etc., thereby increasing the diversity of root verbs in Chinese. The word "写"(write) was used only 26 times.

\begin{table*}[t]
  \centering
  \small
  \begin{tabular}{l r rr rr rr rr rr rr}
    \toprule
    \multirow{2}{*}{\bf Model} & \multicolumn{1}{c}{\bf En} & \multicolumn{2}{c}{\bf Zh} & \multicolumn{2}{c}{\bf Es} & \multicolumn{2}{c}{\bf Ru} & \multicolumn{2}{c}{\bf Fr} & \multicolumn{2}{c}{\bf Ar} & \multicolumn{2}{c}{\bf Avg.} \\
    \cmidrule(lr){2-2} \cmidrule(lr){3-4} \cmidrule(lr){5-6} \cmidrule(lr){7-8} \cmidrule(lr){9-10} \cmidrule(lr){11-12} \cmidrule(lr){13-14}
     & F. & F. & L. & F. & L. & F. & L. & F. & L. & F. & L. & F. & L. \\
    \cmidrule(lr){1-1} \cmidrule(lr){2-2} \cmidrule(lr){3-4} \cmidrule(lr){5-6} \cmidrule(lr){7-8} \cmidrule(lr){9-10} \cmidrule(lr){11-12} \cmidrule(lr){13-14}
     GPT-4  & \bf 58.9 & \bf 58.3 & \bf 62.8  & \bf 54.2 & \bf 47.2 & \bf 52.1 & \bf 51.7  & \bf 52.4 & \bf 54.6  & \bf 57.2 & \bf 56.7  & \bf 55.5 & \bf 54.6 \\
     GPT-3.5-Turbo & 50.0 & 50.0 & 50.0 & 50.0 & 50.0 & 50.0 & 50.0  & 50.0 & 50.0  & 50.0 & 50.0  & 50.0 & 50.0  \\
     Guanaco-13b  & 29.0 & 8.6 & 9.1 & 16.9 & 19.3 & 15.4 & 23.4 & 17.3 & 14.7 & 1.6 & 1.9 & 14.8 & 13.7 \\
     Chimera-I-Chat-13b   & 15.5 & 9.7 & 8.2 & 11.8 & 12.2 & 13.7 & 16.0 & 13.8 & 14.2 & 2.6 & 3.9 & 12.9 & 10.9 \\
     Phoenix-I-Chat-7b  & 6.9 & 13.3 & 21.7 & 7.4 & 8.7 & 2.9 & 4.7 & 8.1 & 8.2 & 8.9 & 11.1 & 7.9 & 10.8 \\
     PolyLM-Chat-13b & 7.7 & 14.0 & 21.2 & 6.1 & 7.6 & 5.5 & 8.3 & 4.8 & 5.1 & 5.5 & 8.5 & 7.3 & 10.2 \\
     Okapi-7b  & 8.8 & 6.2 & 6.9 & 5.0 & 6.1 & 12.1 & 18.9 & 8.7 & 8.7 & 0.3 & 0.0 & 6.8 & 8.1 \\
     PolyLM-MA-13b  & 3.4 & 5.0 & 6.9 & 2.1 & 2.5 & 5.1 & 9.4 & 2.2 & 3.1 & 5.0 & 9.0 & 3.8 & 6.2 \\
     Guanaco-7b  & 4.6 & 3.8 & 5.2 & 0.4 & 0.0 & 1.8 & 3.6 & 1.2 & 2.0 & 0.7 & 0.5 & 2.1 & 2.3 \\
     Bloomz-7b1-mt  & 0.0 & 0.9 & 0.9 & 0.1 & 0.5 & 0.5 & 1.1 & 0.3 & 0.0 & 0.6 & 0.9 & 0.4 & 0.7 \\
     \bottomrule
  \end{tabular}
\caption{Evaluation results of representative multilingual LLMs on OMGEval. ``F.'' denotes the full test set, while ``L.'' represents the localization subset that mainly consists of questions closely-related to a specific language.}
\label{tab:result_1}
\end{table*}

\section{Experiments}

\subsection{Setup}
Initially, we opted for the GPT-text-davinci-003 model as the baseline for evaluating model rankings. However, due to OpenAI shutting down the GPT-text-davinci-003 model on January 4, 2024\footnote{https://platform.openai.com/docs/deprecations}, we substituted it with the GPT-3.5-turbo model as the new baseline for our evaluation.

\subsection{Models}
To evaluate LLMs' capabilities in handling multilingual open-ended Q\&A tasks, we mainly research ten well-known models, organizing them into two groups for comparative analysis. The first group consists of the proprietary GPT family models, renowned for their robust performance across diverse tasks. The second group encompasses open-source, multilingual models, which are vital for understanding the broader landscape of language model proficiencies in various linguistic contexts.

Here we focus on introducing existing open-source multilingual models. BLOOMZ ~\cite{muennighoff2022crosslingual} emerges as a significant multilingual LLM by BigScience, capable of text generation in 46 languages and 13 programming languages, and has been fine-tuned with a multilingual task mixture for diverse linguistic tasks. In parallel, PolyLM ~\cite{wei2023polylm} stands out by addressing the gaps in current models, with proficiency in 18 major languages and enhanced instruction-following capabilities for non-native English instructions. Okapi ~\cite{lai2023okapi} represents the first attempt at instruction-tuned, reinforcement-learning-based multilingual LLMs, broadening the horizon with data in 26 languages. Phoenix ~\cite{phoenix-2023} claims the pioneering status as the first open-source multilingual ChatGPT, mainly focused on non-Latin languages. Guanaco\footnote{https://guanaco-model.github.io/} builds upon Meta's LLaMA 7B model ~\cite{touvron2023llama}, integrating an expansive dataset to excel in multilingual and grammatical tasks across languages like English, Chinese, and German. Lastly, Chimera specializes in Latin languages, symbolizing the fusion of diverse cultural heritages and the democratization of ChatGPT. Together, these models reflect the cutting-edge of polyglot AI, pushing the boundaries of language technology.

\subsection{Results}

\paragraph{GPT v.s. Open-source Models}
As shown in Table ~\ref{tab:result_1},the contrast in performance between the GPT series and other open-source models in our analysis reveals critical insights into the current state of language model development. GPT-4, with their superior resources and unique training methods, set a high bar, exhibiting robust multilingual capabilities that are yet to be matched by open-source alternatives. Interestingly, while GPT-4's dominance is evident, reflecting the culmination of iterative improvements and extensive data training, the strength of GPT-3.5-turbo also suggests that even slightly older iterations of GPT models remain highly competitive. Among the open-source multilingo models, Guanaco-13b performes the best. Furthermore, Guanaco-13b and Chimera-Inst-Chat-13b have a slight advantage. Bloomz's performance was poor, with a win rate of no more than 1 in every language. Overall，the performance of other open-source multilingual models reveals a significant gap in the ability to process and understand cultural nuances, indicating a broader issue within the field. This disparity underscores the critical need for a concerted effort in the community to address cultural biases and enhance the global applicability of LLMs. Figure ~\ref{tab:result_1} shows a case where the answers of three open-source multi-language models have some factual errors for a question in the Chinese localization subset, while GPT-4's answer is right.

 %The results show that GPT-4 is the only model that surpasses the 50 average win rate. However, its 55.52 win rate indicates that OMGEval is a challenging benchmark for current LLMs. In contrast, the performance of other open-source multilingual models reveals a significant gap in the ability to process and understand cultural nuances, indicating a broader issue within the field. This disparity underscores the critical need for a concerted effort in the community to address cultural biases and enhance the global applicability of LLMs. 
 
\begin{figure*}[ht]
    \centering
    \includegraphics[width=0.9\linewidth]{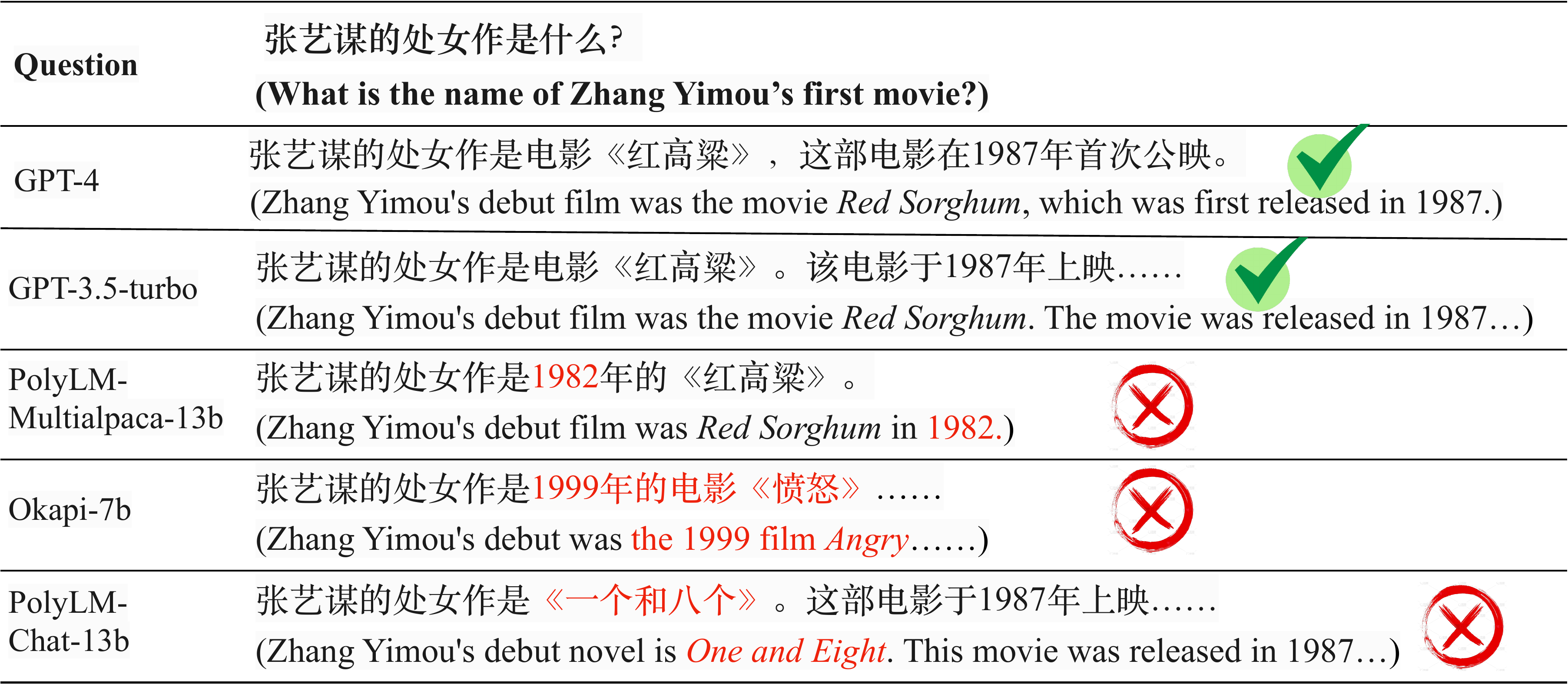}
    \caption{Example of outputs generated by different models, where the answers of GPT-4 and GPT-3.5-turbo are correct and the answers from the other three open-source models are wrong.}
    \label{fig:case}
\end{figure*}
\paragraph{Localization Subset Result}
 As shown in Table ~\ref{tab:result_1}, the ranking remains unchanged compared to the full dataset. However, ther is a slight decline for Guanaco-13b, Chimera-Inst-Chat-13b and Phoenix-Inst-Chat-7b. And the other models show a slight increase. Overall, there is significant room for improvement in the performance of open-source models.

\subsection{Co-relation with human evaluation}
\begin{table}[ht]
    \centering
    \small
    \setlength{\tabcolsep}{8pt} % Increase the padding around the text within each cell
    \begin{tabular}{l c c c c}
    % {
    % >{\centering\arraybackslash}m{2cm} 
    %                 >{\centering\arraybackslash}m{1cm} 
    %                 >{\centering\arraybackslash}m{1cm} 
    %                 >{\centering\arraybackslash}m{1cm}}
    \toprule
    \bf Language & \bf Testset & \bf{Presion} & \bf{Recall} & \bf{F1} \\
    \midrule
    \multirow{2}{*}{\bf Chinese} & \bf F. & 0.81 & 0.94 & 0.88 \\
    & \bf L. & 0.70 & 0.91 & 0.79 \\
    \cmidrule(lr){1-2} \cmidrule(lr){3-5}
    \multirow{2}{*}{\bf Spanish} & \bf F. & 0.92 & 0.92 & 0.91 \\
    & \bf L. & 0.90 & 0.90 & 0.89 \\
    % \bf{Es} & 0.92 & 0.92 & 0.91 \\
    % \bf{Es-local} & 0.90 & 0.90 & 0.89 \\ \hline
    \bottomrule
    \end{tabular}
    \caption{Presion, Recall and F1 between the evaluation from human and GPT-4. We provide the answers from GPT-3.5-Turbo and PolyLM-chat-13B, and ask the evaluators (i.e., humans or GPT-4) to select the better one.}
    \label{tab:human_ann}
\end{table}

AlpacaEval has demonstrated consistency in English, with GPT-4 as an evaluator alongside human labeling. However, our primary interest is the model's performance on multilingual tasks. We conducted human annotations for two languages, Chinese and Spanish, curating a sample of 100 data points for each language, comprising 50 localized and 50 other data. Responses from two models, GPT-3.5-Turbo and PolyLM-chat-13B, were selected, and human judges were tasked with assessing which model delivered superior responses regarding grammatical accuracy, relevance, informativeness, and factual correctness. Judges could either select one model as the winner or indicate no significant difference between the models. Our Chinese annotators are native speakers holding master's degrees in linguistics, while the Spanish annotators are students with a minimum level 4 proficiency in Spanish. The accompanying Table~\ref{tab:human_ann} illustrates that GPT-4's scores align closely with human annotations, affirming the robustness of GPT-4 for evaluating multilingual capabilities.

\section{Related Work}
 We will explore related work through two distinct lenses: multilingual evaluation and generative evaluation.

\paragraph{Multilingual Evaluation}
The multilingual ability of LLMs is a significant dimension of interest. High-quality datasets offer an objective and thorough evaluation of this proficiency, aiding researchers in developing more impressive models. On Nature Language Understanding, XNLI ~\cite{conneau2019unsupervised} is a benchmark for appraising the model's proficiency in grasping textual meanings. Similarly, PAWS-X ~\cite{yang2019paws} challenges the model to discern whether a sentence is a paraphrase of another. XCOPA ~\cite{ponti2020xcopa}, on the other hand, engages the model in making inferential judgments based on given premises. XWinograd ~\cite{tikhonov2021s} tests the model's ability to identify pronoun references within a sentence accurately. Additionally, Belebele ~\cite{bandarkar2023belebele} stands out as a typical reading comprehension dataset structured in a multiple-choice format. Open Domain Q\&A, pioneered by ~\citet{green1961baseball}, is also enriched with various multilingual resources such as XQA ~\cite{liu2019xqa}, TyDi-QA ~\cite{clark2020tydi}, and Xor-QA ~\cite{asai2020xor}. Moreover, datasets like MLQA ~\cite{lewis2019mlqa}, XQuAD ~\cite{artetxe2019cross}, and MKQA ~\cite{longpre2021mkqa} offer parallel questions across languages. Interestingly, LLMs have the opportunity to demonstrate their intelligence by undertaking human examinations, as exemplified by the M3Exam ~\cite{zhang2023m3exam}. 

\paragraph{Generative Evaluation}
As humans communicate through speech or writing, LLMs similarly express themselves by generating. The popularity of ChatGPT has spurred the development of generative models. Traditional evaluations of these models often revolved around restricted question-and-answer formats with definitive answers, typical in NLP tasks like Grammatical Error Correction ~\cite{ng2014conll, bryant2019bea}, Chinese Spelling Correction ~\cite{yu2014overview, hu2022cscd}. However, such narrow evaluation only partially captures a model's generative capabilities, limiting its potential to a constrained framework. More open-ended evaluation benchmarks have been introduced in response to this limitation. For instance, the Stanford AlpacaEval is widely recognized in English, while SuperCLUE ~\cite{xu2023superclue} serves a similar purpose in Chinese. In these evaluations, the outputs generated by models are assessed by experts or specialized models for quality and relevance. Moreover, MT-Bench ~\cite{zheng2023judging} has been developed to evaluate model performance in sustained, multi-turn dialogues featuring prompts that necessitate multiple rounds of interaction.

\section{Conclusions}

In this work, we propose an open multilingual generative evaluation benchmark for LLMs, which can provide an automatic quantitative evaluation for LLMs in diverse cultural contexts. We evaluate several multilingual LLMs, including both closed-source and open-source ones, on OMGEval. 
% The results indicate that open-source multilingual LLMs are significantly weaker than closed-source ones, 
% Experiments demonstrate that there is still significant room for current multilingual models. 

\section{Ethical Impact}
The annotators and reviewers involved in the construction of our dataset were remunerated accordingly, with a total compensation of approximately 2,500 dollars. We highly recommend users to utilize our work exclusively for research purposes, with the objective of enhancing the ability of LLMs across various cultural background.

\section{Limitations}
In the paper, our proposed dataset only contains five language types with various culture backgrounds. Nevertheless, the method can be easily extended to other languages. We leave it to the future work to include more languages.

\bibliography{custom}

\appendix

\end{CJK*}
\end{document}